\documentclass{article}
\usepackage[T1]{fontenc}
\usepackage{iclr2027_conference,times}
\usepackage{amsmath,amssymb,amsthm,bm,mathtools}
\usepackage{graphicx,booktabs,multirow,tabularx}
\usepackage{algorithm,algpseudocode}
\usepackage{microtype,xcolor,hyperref,url,placeins}
\hypersetup{colorlinks=true,citecolor=teal,linkcolor=teal,urlcolor=teal,pdftitle={Kernel Ridge Attention: Nonlinear Memory with Selective Forgetting},pdfauthor={Haowei Chen, Yichen Wang, and Congfu Xu}}
\iclrfinalcopy
\newcommand{\R}{\mathbb{R}}
\newcommand{\I}{\mathbf{I}}
\newcommand{\tr}{\operatorname{tr}}
\newcommand{\diag}{\operatorname{diag}}
\newcommand{\softplus}{\operatorname{softplus}}
\newcommand{\silu}{\operatorname{SiLU}}

\makeatletter
\let\inputtable\@@input
\makeatother
\newtheorem{proposition}{Proposition}

\title{Kernel Ridge Attention:\\Nonlinear Memory with Selective Forgetting}
\author{Haowei Chen\\Independent Researcher\\{\footnotesize\texttt{chenhaowei0019@gmail.com}}
\And Yichen Wang\\Cornell University\\{\footnotesize\texttt{yw2598@cornell.edu}}
\And Congfu Xu\thanks{Corresponding author.}\\Zhejiang University\\{\footnotesize\texttt{xucongfu@zju.edu.cn}}}
\begin{document}
\maketitle
\lhead{Preprint}
\begin{abstract}
Kernel Ridge Attention (KRA) is a recurrent sequence mixer that combines learned nonlinear features with covariance-guided memory updates. It stores associations through recursive ridge regression, using correlations among past keys to determine how new observations modify memory. A content-dependent Ornstein--Uhlenbeck transition controls retention and restores uncertainty for future writes. This construction admits an exact temporal-kernel interpretation and maintains an inference state whose size is independent of context length. Associative block maps and explicit adjoints support parallel training while preserving the recursive update. In controlled language-modeling experiments, KRA improves prediction quality over attention and recurrent baselines, strengthens natural associative recall over the evaluated recurrent models, and achieves practical distributed training throughput. These properties provide a principled foundation for nonlinear, adaptive sequence memory.
\end{abstract}

\section{Introduction}
Efficient sequence models must repeatedly decide what to retain, what to overwrite, and how to retrieve it. Softmax attention preserves a separate key and value for every past token. Recurrent alternatives write these associations into a fixed-size state \citep{katharopoulos2020linear,dao2024ssm}. This changes the central modeling problem: memory updates must manage interference among associations that share the same state.

The delta rule addresses interference by writing a prediction residual. Gated DeltaNet additionally controls retention with an input-dependent decay \citep{yang2025gated}. These mechanisms make recurrent memory selective in both writing and forgetting. A further question concerns the \emph{geometry} of a write: when past keys are correlated, which directions should a new observation modify? A scalar write strength treats those directions uniformly. Ridge regression provides a different answer by using the inverse regularized key covariance to shape the update.

We develop this idea into \textbf{Kernel Ridge Attention (KRA)}. Each head constructs unit-norm nonlinear Fourier features and maintains a regression matrix together with its posterior covariance. The covariance transforms a key into a write direction, suppressing directions already constrained by previous observations. A learned content clock evolves this memory through an Ornstein--Uhlenbeck (OU) process. The transition both attenuates old coefficients and replenishes covariance toward a stationary prior, coupling selective forgetting with the capacity to learn new associations.

This construction has a direct kernel interpretation. Conditioned on the realized feature and clock sequence, the recurrent output equals prefix kernel ridge regression with a product of a learned content kernel and an exact exponential time kernel. The content representation has finite dimension, while the temporal kernel is implemented through Markov state transitions. Consequently, the model retains a constant-size inference state through Markov transitions that summarize the prefix.

The architecture applies short causal convolutions, projections, and SiLU to the query, key, and value paths. Query and key then enter a shared learned Fourier map. We express the covariance evolution as a composition of block maps and the coefficient evolution as an affine recurrence. Associative prefix scans compute block boundaries, and explicit joint adjoints provide the corresponding backward computation. This organization preserves recursive least squares while exposing parallelism across sequence blocks.

We evaluate KRA using a shared backbone, token stream, and training budget. At roughly 31M parameters and 100M training tokens, KRA reaches 88.00 perplexity, improving over Transformer++ by 9.47\% and Gated DeltaNet by 20.68\%. Natural rare-repeat top-1 accuracy reaches 25.41\%, compared with 6.69\% for Gated DeltaNet. The optimized eight-GPU training step takes 31.77\,ms, corresponding to 1.03M tokens/s. The recurrent inference state occupies 1.165\,MiB at both 4K and 8K prefixes.

Our contributions are threefold: (1) a nonlinear recurrent attention layer with covariance-guided writes and selective OU forgetting; (2) an exact prefix-kernel interpretation with associative block maps and explicit joint adjoints; and (3) a controlled evaluation of language modeling, associative recall, feature components, and distributed training efficiency.

\begin{figure}[t]
\centering
\includegraphics[width=\linewidth]{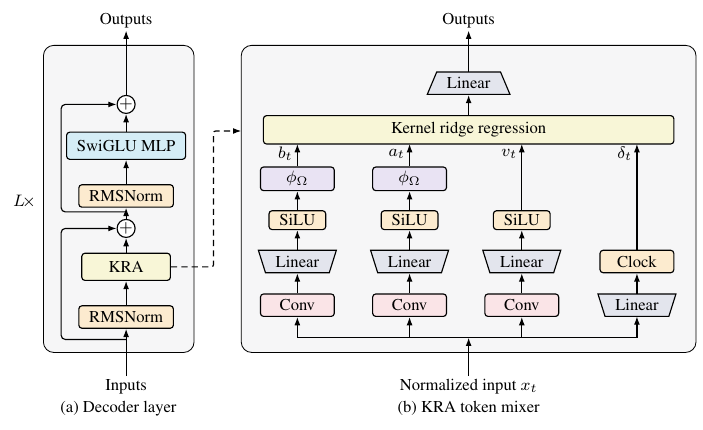}
\caption{\textbf{Kernel Ridge Attention.} Left: a pre-norm decoder layer with KRA and a SwiGLU MLP. Right: the expanded mixer. Query, key, and value paths apply causal convolution, projection, and SiLU. Query and key additionally share a learned Fourier map. The content clock controls OU prediction. Each head carries covariance $P$ and regression coefficients $\Theta$, writes the current value, and reads with the query.}
\label{fig:architecture}
\end{figure}

\section{Kernel Ridge Attention}
\label{sec:method}
We describe a single head; the layer applies the construction independently across heads. The head dimension and value dimension are $d=d_v=64$. The content feature dimension is $D=2m=64$, with $m=32$ learned frequencies. Vectors are columns, and $x_t$ denotes the normalized layer input.

\subsection{Learned content features}
\label{sec:features}
Let $\mathcal C_Q$, $\mathcal C_K$, and $\mathcal C_V$ be independent causal depthwise convolutions of width four. Each path extracts local input-channel patterns before projection and SiLU. The value is $v_t=\silu\!\left(W_V(\mathcal C_Vx)_t\right)$, and the query and key are
\begin{equation}
q_t=\silu\!\left(W_Q(\mathcal C_Qx)_t\right),\qquad
k_t=\silu\!\left(W_K(\mathcal C_Kx)_t\right).
\label{eq:frontend}
\end{equation}
For a frequency matrix $\Omega\in\R^{m\times d}$, define
\begin{equation}
\phi_\Omega(z)=\frac{1}{\sqrt m}
\begin{bmatrix}\cos(\Omega z)\\\sin(\Omega z)\end{bmatrix},\qquad
a_t=\phi_\Omega(k_t),\quad b_t=\phi_\Omega(q_t).
\label{eq:features}
\end{equation}
The map has unit norm and induces the positive semidefinite kernel
\begin{equation}
\kappa_\Omega(z,z')=\phi_\Omega(z)^\top\phi_\Omega(z')
=\frac1m\sum_{r=1}^{m}\cos\!\left(\omega_r^\top(z-z')\right).
\label{eq:contentkernel}
\end{equation}
Query and key share the bank within a head; different heads and layers have independent banks. We initialize orthogonal directions with Gaussian QR and use scrambled Sobol samples for the radial $\chi_d$ quantiles, following the motivation of orthogonal random features \citep{rahimi2007random,yu2016orthogonal}. The initial spectral scale is $\beta_0=1/8$. During training, both the bank and a positive bandwidth offset are learned: $\Omega_{\mathrm{eff}}=\exp(\ell_\beta/2)\Omega_{\mathrm{param}}$, where the initial bank already contains $\sqrt{\beta_0}$. Equation~\eqref{eq:contentkernel} defines the learned kernel exactly.

\subsection{Selective time and uncertainty recovery}
Each head computes a positive content-dependent step and a retention factor,
\begin{equation}
\rho_t=\exp(A_{\log})\softplus(w_a^\top x_t+b_{\Delta t}),\qquad
\delta_t=\exp(-\rho_t),\qquad u_t=\sum_{j=1}^{t}\rho_j.
\label{eq:clock}
\end{equation}
The cumulative coordinate $u_t$ supports the kernel interpretation, and the recurrent transition uses the current $\delta_t$. A learned positive ridge $\lambda=\exp(\ell_\lambda)$ is initialized to one. The recurrent state consists of $\Theta_t\in\R^{D\times d_v}$ and $P_t\in\R^{D\times D}$, initialized as $\Theta_0=0$ and $P_0=\lambda^{-1}\I$. Before observing the current value, the model predicts
\begin{equation}
\bar\Theta_t=\delta_t\Theta_{t-1},\qquad
\bar P_t=\delta_t^2P_{t-1}+(1-\delta_t^2)\lambda^{-1}\I.
\label{eq:ou}
\end{equation}
These are the exact mean and covariance transitions of a stationary OU process sampled at the learned steps. The second term restores uncertainty toward the prior: forgetting old evidence also reopens directions for future writes.

\subsection{Covariance-guided writing}
The observation update uses the residual under the predicted regression matrix:
\begin{align}
z_t&=\bar P_ta_t,&s_t&=1+a_t^\top z_t,&G_t&=z_t/s_t,\nonumber\\
e_t&=v_t-\bar\Theta_t^\top a_t,&
\Theta_t&=\bar\Theta_t+G_te_t^\top,&
P_t&=\bar P_t-z_tz_t^\top/s_t.
\label{eq:update}
\end{align}
We read after writing, $o_t=\Theta_t^\top b_t$. This inclusive update predicts the next token from the current causal prefix. Each value channel has its own regression coefficients and shares the same covariance.

The update is the unique solution of a local regularized regression problem:
\begin{equation}
\Theta_t=\arg\min_\Theta\left\{
\frac12\|v_t-\Theta^\top a_t\|_2^2+
\frac12\tr\!\left[(\Theta-\bar\Theta_t)^\top\bar P_t^{-1}(\Theta-\bar\Theta_t)\right]
\right\}.
\label{eq:localobjective}
\end{equation}
Thus $\bar P_t^{-1}$ supplies the metric of the write. In the isotropic case $\bar P_t=c\I$, $G_t=c a_t/(1+c\|a_t\|^2)$ is a scalar multiple of the key. The full covariance rotates and rescales this direction according to past observations. This is the structural distinction from a scalar-gated residual write.

\begin{proposition}[Bounded covariance and writing]
For $\lambda>0$ and $0<\delta_t\leq1$, the recurrence satisfies $0\prec P_t\preceq\lambda^{-1}\I$. For unit-norm features, the self-write coefficient obeys
\begin{equation}
0<\frac{a_t^\top\bar P_ta_t}{1+a_t^\top\bar P_ta_t}\leq\frac{1}{1+\lambda}.
\label{eq:writebound}
\end{equation}
\end{proposition}
The current value's coefficient in the query output is $b_t^\top G_t$, which depends on the query--key alignment. This separates the capacity to write an observation from the ability of a particular query to retrieve it.

\subsection{Exact prefix kernel regression}
\label{sec:dual}
Fix a realized sequence of features and clock values. On the prefix $1{:}t$, define the key Gram matrix and query cross-covariance
\begin{equation}
C_{ij}=a_i^\top a_j\,e^{-|u_i-u_j|},\qquad
d_{t,i}=b_t^\top a_i\,e^{-(u_t-u_i)}.
\label{eq:dualkernel}
\end{equation}
\begin{proposition}[Recurrent--dual equivalence]
Let $V_{1:t}$ stack the observed value vectors as rows. The output of Equations~\eqref{eq:ou}--\eqref{eq:update} equals
\begin{equation}
o_t^\top=d_t^\top(C_{1:t,1:t}+\lambda\I)^{-1}V_{1:t}.
\label{eq:dual}
\end{equation}
\end{proposition}
The content Gram and OU time Gram are positive semidefinite, so their entrywise product $C$ is positive semidefinite. Positive ridge makes the solve strictly positive definite. The time factor is exact: a finite Markov state realizes its exponential covariance. This uses the state-space interpretation of temporal Gaussian processes \citep{hartikainen2010kalman,sarkka2012spatiotemporal}.

The regression system is built from keys; queries evaluate the learned function at new feature points. Distinct $W_Q$ and $W_K$ therefore fit naturally into this construction. Positive definiteness belongs to the feature-domain kernel and its key Gram, while the query--key cross-covariance supplies the readout. The kernel identity is conditional on the realized layer-input path, including its content-dependent clock. Appendix~\ref{app:proofs} gives the proofs and the corresponding dynamic least-squares objective.

\section{Parallel Recursive Computation}
\label{sec:engineering}
KRA's token update uses $O(D^2+Dd_v)$ arithmetic and carries $D^2+Dd_v$ scalars per head. We organize a sequence into blocks and precompute their state transformations. The resulting composition laws expose parallelism in both the forward recurrence and its adjoints.

\subsection{Covariance and coefficient prefix maps}
A block acts on an incoming covariance through a Riccati map
\begin{equation}
\Phi(P)=C+A(P^{-1}+J)^{-1}A^\top,
\label{eq:riccati-map}
\end{equation}
where $C,J\succeq0$. The block statistics $(A,C,J)$ follow from local Sherman--Morrison updates and incorporate the OU process noise. These maps are closed under composition, following the associative filtering formulation of \citet{sarkka2019parallel}. For an earlier block $l$ and a later block $r$, let $N=(\I+C_lJ_r)^{-1}$. The composed map has
\begin{align}
A_{r\circ l}&=A_rNA_l,\nonumber\\
C_{r\circ l}&=C_r+A_rNC_lA_r^\top,\nonumber\\
J_{r\circ l}&=J_l+A_l^\top J_rNA_l.
\label{eq:riccati-compose}
\end{align}
Associativity supplies all block-boundary covariances through a prefix scan. Each block then evaluates Equation~\eqref{eq:update} from its boundary to obtain the gains $G_i$.

Given these gains, write $T_i=\delta_i(\I-G_i a_i^\top)$. The coefficient recurrence is affine:
\begin{equation}
\Theta_i=T_i\Theta_{i-1}+G_iv_i^\top,
\qquad \Theta_{\rm out}=F\Theta_{\rm in}+B.
\label{eq:affine-map}
\end{equation}
Two blocks compose as $(F_rF_l,F_rB_l+B_r)$, giving a second prefix scan. The within-block outputs can also be written as causal matrix products. With $c_i=\prod_{j\leq i}\delta_j$ and the innovations $e_j$ from Equation~\eqref{eq:update},
\begin{equation}
o_i^\top=c_i b_i^\top\Theta_{\rm in}
+\sum_{j\leq i}\frac{c_i}{c_j}(b_i^\top G_j)e_j^\top.
\label{eq:causal-read}
\end{equation}
Decay ratios are evaluated from sums of log retention. Across blocks, the global dependency is organized into logarithmically many scan stages; each block retains its short recursive update.

\subsection{Explicit joint adjoints}
Let $D=\partial\mathcal L/\partial\Theta$ and let $S$ be the symmetric covariance adjoint, $S=\partial\mathcal L/\partial P+(\partial\mathcal L/\partial P)^\top$. Define $F_i=T_i\cdots T_1$, $F_0=\I$, and the block statistic
\begin{equation}
w_i=\delta_i F_{i-1}^\top a_i,
\qquad E=\sum_i w_i e_i^\top/s_i.
\label{eq:cross-statistic}
\end{equation}
Perturbations of the incoming states obey
\begin{equation}
\Delta P_{\rm out}=F\Delta P_{\rm in}F^\top,\qquad
\Delta\Theta_{\rm out}=F\Delta\Theta_{\rm in}+F\Delta P_{\rm in}E.
\end{equation}
Consequently, the block backward map is
\begin{align}
D_{\rm in}&=F^\top D_{\rm out}+D_{\rm local},\nonumber\\
S_{\rm in}&=F^\top S_{\rm out}F+F^\top D_{\rm out}E^\top
+ED_{\rm out}^\top F+S_{\rm local}.
\label{eq:joint-adjoint}
\end{align}
The local source terms collect the block's output gradients. Causal block products compute these terms in parallel, and a reverse prefix scan supplies the incoming adjoints for each local backward recurrence. Appendix~\ref{app:block} derives the statistics and the explicit token adjoints.

\subsection{Working-precision evaluation}
The covariance maps require fixed-size state-space inverses. Their positive spectral structure provides bounds for polynomial inverse evaluation; a recomputed residual $R=\I-WX$ then corrects an approximate inverse through $X\leftarrow X+XR$. The stationary prior also simplifies the backward computation: for $P_0=r\I$ with $r=\lambda^{-1}$, its initial-state contribution is $\tfrac12\tr(S_0)$.

We use 64-token blocks, BF16 network activations, and FP32 features, recurrent states, and adjoints. Block matrix products use compensated arithmetic and residual correction. The implementation follows the explicit recurrences above, with saved boundary states and local recomputation. Numerical validation covers outputs, all state and input gradients, causality, and streaming continuation.

\section{Experiments}
\label{sec:experiments}
We study language-modeling quality, associative recall, training efficiency, and the learned content features. All reported models use the same training-token stream and final validation windows.

\subsection{Experimental setup}
\textbf{Models and training.} The common decoder has six layers, width 384, six heads, a head dimension of 64, and a SwiGLU feed-forward width of 1152. It uses pre-RMSNorm residual blocks and a tied input/output embedding with 50,257 tokens. KRA has 30.92M parameters. We compare it with Transformer++, Gated DeltaNet, and two recurrence controls that replace the Gated DeltaNet core with KDA or GDN-2 updates \citep{kimiteam2025linear,hatamizadeh2026gdn2}. The latter controls preserve the Gated DeltaNet outer layer, use direct full-rank gate projections, and initialize the expanded gates to reproduce its scalar-gate function. Their parameter counts are reported explicitly.

Each run trains for 3,052 updates on 100,007,936 prediction tokens from a fixed SlimPajama sample \citep{cerebras2023slimpajama}, tokenized with GPT-2. The global batch is 16 sequences of length 2,048, with microbatch two on each of eight RTX 5090 GPUs. We use AdamW, a peak learning rate of $10^{-3}$, 100 warmup steps, a cosine decay to $10^{-4}$, weight decay 0.1, and gradient clipping at one. The last checkpoint is evaluated on 512 fixed windows containing 1,048,576 prediction tokens. Curves use a shared subset of 128 windows. All runs use one training seed and one shared data order; Appendix~\ref{app:setup} records initialization, data construction, and measurement details.

\textbf{Recall.} Following the focus on natural associative recall in \citet{arora2024zoology}, we identify repeated token bigrams within the same document and window, selecting bigrams seen at most twelve times in the training stream. This yields 31,319 rare-repeat targets, including 2,749 with a previous occurrence at distance at least 512. We also use 3,584 controlled prompts that balance two target values, matched and mismatched cues, and seven cue--target distances. We report both balanced choice accuracy and association gain, a log-probability contrast defined in Appendix~\ref{app:recall}.

\subsection{Language modeling and recall}
\begin{table}[htbp]
\centering
\caption{\textbf{Language modeling at a shared 100M-token budget.} KDA and GDN-2 are recurrence controls in the Gated DeltaNet backbone. Lower NLL and perplexity are better.}
\label{tab:main}
\small
\begin{tabular}{lrrr}
\toprule
Model & Params (M) & NLL & PPL\\
\midrule
\inputtable tables/main.tex
\bottomrule
\end{tabular}
\end{table}

\begin{figure}[t]
\centering
\includegraphics[width=\linewidth]{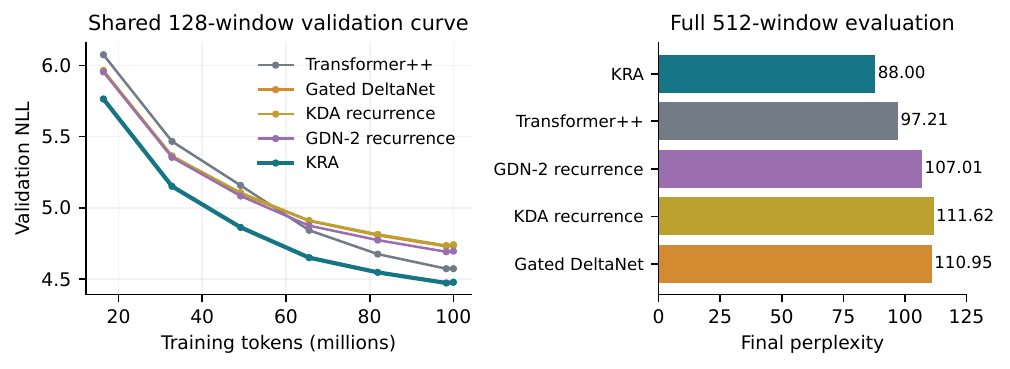}
\caption{\textbf{Language-modeling comparison.} Left: NLL on the shared curve subset, starting at the first evaluation after initialization. Right: perplexity of the final checkpoint on the complete validation stream. Curves and final values use different, explicitly fixed evaluation sizes.}
\label{fig:quality}
\end{figure}

KRA reaches a validation NLL of 4.4774 and perplexity of 88.00 (Table~\ref{tab:main}). Perplexity improves by 9.47\% over Transformer++, 20.68\% over Gated DeltaNet, and 17.76\% over the GDN-2 recurrence control. Figure~\ref{fig:quality} shows the validation trajectories and final checkpoint comparison.

\begin{table}[htbp]
\centering
\caption{\textbf{Recall within 2K-token windows.} Rare-repeat top-1 and controlled-copy balanced accuracy are percentages. ``Far'' denotes distance $\geq512$; copy gain is measured in nats.}
\label{tab:recall}
\small
\begin{tabular}{lrrrrr}
\toprule
Model & Rare top-1 & Far top-1 & Far NLL & Copy gain & Copy acc.\\
\midrule
\inputtable tables/recall.tex
\bottomrule
\end{tabular}
\end{table}

\begin{figure}[t]
\centering
\includegraphics[width=\linewidth]{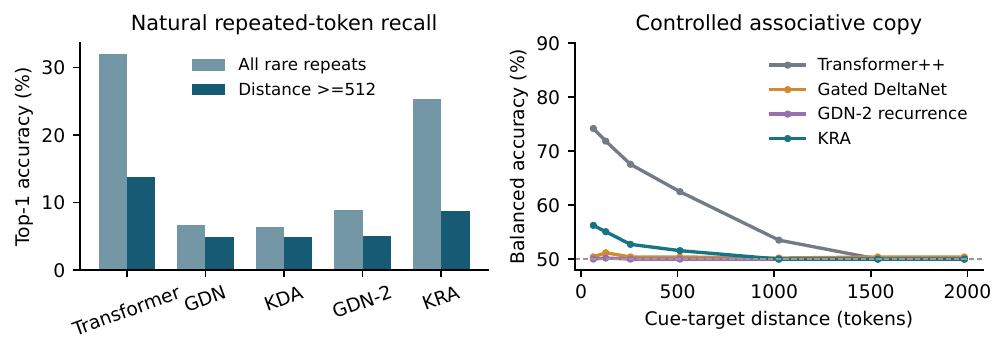}
\caption{\textbf{Natural and controlled recall.} The natural slice measures repeated content in held-out text. The controlled probe varies cue--target distance while balancing answer and cue identity; its dashed line marks chance accuracy.}
\label{fig:recall}
\end{figure}

On natural rare repeats, KRA attains 25.41\% top-1 accuracy, compared with 6.69\% for Gated DeltaNet and 8.95\% for the GDN-2 recurrence control (Table~\ref{tab:recall}). KRA reaches 8.69\% top-1 on rare repeats at distance at least 512. Transformer++ reaches 32.00\% on the full rare-repeat slice and the highest scores on distant targets. The controlled copy probe in Figure~\ref{fig:recall} measures association gain and balanced answer choice across cue distances. Together, the natural and controlled probes characterize the retrieval behavior alongside overall language-modeling quality.

\subsection{Core feature ablations}
\label{sec:ablations}
\textbf{Learning the Fourier bank.} A matched bank comparison uses SiLU content features with pre-projection Q/K convolutions and a linear value path with post-projection convolution. Learning the spectral directions yields 92.36 perplexity and 25.20\% rare-repeat top-1, compared with 92.96 and 23.60\% for a fixed initial bank (Table~\ref{tab:coreablations}). Both arms share their initial function and learn the bandwidth. Bank learning adds 73,728 trainable parameters, or 0.24\% of the model.

\textbf{The activation before the Fourier map.} The Q/K activation pair shares post-projection convolutions in both arms. SiLU yields 94.95 perplexity and 20.46\% rare-repeat top-1; the identity yields 96.56 and 18.06\%. The Fourier bank, bandwidth, ridge, and content clock remain learned throughout.

\textbf{Value activation.} With pre-projection convolutions on all three paths, SiLU values yield 88.00 perplexity, compared with 93.16 for linear values. Far-repeat top-1 rises from 8.00\% to 8.69\%. These matched comparisons connect feature and value construction to prediction and retrieval behavior.

\begin{table}[htbp]
\centering
\caption{\textbf{Core content-feature ablations.} Each group shares its front end, initial tensors, clock, ridge, and bandwidth. The bank and Q/K activation groups use the architectures described in the text; the value-activation group uses pre-projection convolutions on Q/K/V.}
\label{tab:coreablations}
\small
\begin{tabular}{lrrr}
\toprule
Variant & PPL & Rare top-1 (\%) & Far top-1 (\%)\\
\midrule
\inputtable tables/core_ablations.tex
\bottomrule
\end{tabular}
\end{table}

\subsection{Training efficiency and inference state}
\label{sec:cost}
Table~\ref{tab:training} measures the optimized implementations at sequence length 2,048 and microbatch two. KRA completes full-model forward and backward in 21.08\,ms. Its eight-GPU step takes 31.77\,ms, or 1.136 times Gated DeltaNet's 27.97\,ms, and sustains 1.03M prediction tokens/s. The step includes gradient communication, clipping, and the optimizer. Both models share the same backbone and loss optimizations. Appendix~\ref{app:setup} specifies the timing protocol.

\begin{table}[htbp]
\centering
\caption{\textbf{Optimized training latency.} Medians over 50 samples on RTX 5090 GPUs. Full-model forward--backward uses one GPU; complete steps use eight GPUs with global batch 16.}
\label{tab:training}
\small
\begin{tabular}{lrrr}
\toprule
Model & Forward--backward (ms) & Eight-GPU step (ms) & Tokens/s\\
\midrule
\inputtable tables/training.tex
\bottomrule
\end{tabular}
\end{table}

The six-layer KRA inference state occupies 1.165\,MiB at both 4K and 8K prefixes (Table~\ref{tab:cost}). The attention cache occupies 36 and 72\,MiB, respectively; at 8K, KRA uses 61.8 times less persistent storage. The recurrent state holds the covariance, coefficient matrix, and short-convolution histories, and its size is independent of prefix length.

\begin{table}[htbp]
\centering
\caption{\textbf{Persistent inference state.} Batch-one storage after 4K and 8K prefixes, including convolution histories for recurrent models and the key/value cache for attention.}
\label{tab:cost}
\small
\begin{tabular}{lrr}
\toprule
Model & 4K state (MiB) & 8K state (MiB)\\
\midrule
\inputtable tables/cost.tex
\bottomrule
\end{tabular}
\end{table}

\FloatBarrier
\section{Related Work}
\textbf{Recurrent attention and selective updates.} Linear attention converts feature-based attention into recurrent accumulation \citep{katharopoulos2020linear}. DeltaNet uses residual writes and supports hardware-efficient sequence processing \citep{yang2024delta}. Gated DeltaNet combines decay and the delta rule \citep{yang2025gated}; KDA adds channel-wise retention \citep{kimiteam2025linear}, and GDN-2 separates erase and write gates \citep{hatamizadeh2026gdn2}. KRA shares the residual-write structure, with a full covariance determining the write direction and an OU transition restoring its uncertainty.

\textbf{Regression as recurrent computation.} Mesa layers formulate sequence mixing as an inner regression problem \citep{vonoswald2024mesa}, and MesaNet develops an efficient solver-based implementation for language modeling \citep{vonoswald2026mesanet}. Test-Time Training treats the recurrent state as a learned model updated on the current sequence \citep{sun2024ttt}. Preconditioned DeltaNet explicitly incorporates curvature, using diagonal approximations in practical recurrent operators \citep{tumma2026preconditioned}. KRA combines learned nonlinear features with a full covariance and an exact OU product kernel. These components connect adaptive memory geometry to a finite-state realization of nonlinear kernel regression.

\textbf{Kernel features and state-space Gaussian processes.} Random Fourier features approximate stationary kernels with finite-dimensional maps \citep{rahimi2007random}; orthogonal features improve the initialization geometry \citep{yu2016orthogonal}. We use this construction as an initialization and learn the frequencies end to end. Temporal Gaussian processes admit exact state-space representations for kernels with finite Markov realizations \citep{hartikainen2010kalman,sarkka2012spatiotemporal}. KRA uses the OU realization to retain an exact temporal factor while learning a finite content kernel.

Associative filtering and smoothing express temporal inference as parallel prefix operations \citep{sarkka2019parallel}. KRA applies this structure to its covariance and coefficient maps and derives explicit joint adjoints for end-to-end sequence-model training.

\section{Conclusion}
Kernel Ridge Attention combines learned nonlinear features, covariance-guided writes, and a content-dependent OU clock in one recurrent regression layer. Its finite content state realizes an exact temporal kernel. Associative block maps and explicit adjoints organize the recurrence into parallel training computations. In the controlled language-modeling study, KRA improves perplexity over attention and delta-rule baselines and strengthens natural associative recall over the evaluated recurrent models. These results establish nonlinear regression and uncertainty recovery as useful components of fixed-state language-model memory, with a formulation that supports further evaluation across model scales and sequence tasks.
\section*{Acknowledgments}
We thank Jiahui Wang (Alibaba Cloud).
\bibliography{references}
\bibliographystyle{iclr2027_conference}
\clearpage
\appendix
\section{Derivations}
\label{app:proofs}
\subsection{The local regression update}
Differentiating Equation~\eqref{eq:localobjective} with respect to $\Theta$ gives
\begin{equation}
(\bar P_t^{-1}+a_ta_t^\top)\Theta
=\bar P_t^{-1}\bar\Theta_t+a_tv_t^\top.
\end{equation}
Sherman--Morrison gives
\begin{equation}
P_t=(\bar P_t^{-1}+a_ta_t^\top)^{-1}
=\bar P_t-\frac{\bar P_ta_ta_t^\top\bar P_t}{1+a_t^\top\bar P_ta_t}.
\end{equation}
Substitution yields $\Theta_t=\bar\Theta_t+G_t(v_t-\bar\Theta_t^\top a_t)^\top$. The Hessian is positive definite because $\bar P_t$ is positive definite, establishing uniqueness.

\subsection{Covariance and self-write bounds}
Assume $0\prec P_{t-1}\preceq\lambda^{-1}\I$. Equation~\eqref{eq:ou} is a convex combination of $P_{t-1}$ and $\lambda^{-1}\I$, so $0\prec\bar P_t\preceq\lambda^{-1}\I$. The precision update is $P_t^{-1}=\bar P_t^{-1}+a_ta_t^\top$, giving $0\prec P_t\preceq\bar P_t$. Induction from $P_0=\lambda^{-1}\I$ proves the covariance bound. For a unit feature, $h_t=a_t^\top\bar P_ta_t\in(0,\lambda^{-1}]$. Since $h/(1+h)$ is increasing for positive $h$, the self-write coefficient is at most $1/(1+\lambda)$.

\subsection{OU state and the prefix kernel}
For one value channel, consider the stationary process
\begin{equation}
d\theta(u)=-\theta(u)\,du+\sqrt{2/\lambda}\,dW(u),\qquad
\theta(0)\sim\mathcal N(0,\lambda^{-1}\I_D).
\end{equation}
Its exact transition over a positive interval $\rho_t$ is
\begin{equation}
\theta_t=\delta_t\theta_{t-1}+\xi_t,\qquad
\xi_t\sim\mathcal N(0,(1-\delta_t^2)\lambda^{-1}\I_D),
\quad\delta_t=e^{-\rho_t}.
\end{equation}
The stationary covariance is $\operatorname{Cov}(\theta_i,\theta_j)=\lambda^{-1}e^{-|u_i-u_j|}\I_D$. Under observations $v_i=a_i^\top\theta_i+\epsilon_i$ with independent unit-variance noise, the observation covariance is $C/\lambda+\I$. The query $b_t^\top\theta_t$ has cross-covariance $d_t^\top/\lambda$ with those observations. Gaussian conditioning therefore yields
\begin{equation}
\mathbb E[b_t^\top\theta_t\mid v_{1:t}]
=\frac{d_t^\top}{\lambda}\left(\frac{C}{\lambda}+\I\right)^{-1}v_{1:t}
=d_t^\top(C+\lambda\I)^{-1}v_{1:t}.
\end{equation}
The Kalman prediction and observation updates for the same model are exactly Equations~\eqref{eq:ou}--\eqref{eq:update}. Applying the argument to each value channel proves recurrent--dual equivalence.

The positive semidefinite product kernel lives on feature-time points:
\begin{equation}
\kappa((z,u),(z',u'))=\phi_\Omega(z)^\top\phi_\Omega(z')e^{-|u-u'|}.
\end{equation}
Keys provide the regression points $(k_i,u_i)$; the query is $(q_t,u_t)$. This formulation allows separate query and key projections. The key Gram is symmetric and positive semidefinite, and the query cross-covariance evaluates the learned function through its distinct query map. Causality follows by conditioning on each prefix separately.

\subsection{A dynamic least-squares objective}
For $0<\delta_i<1$, the same state is obtained by minimizing over a coefficient trajectory,
\begin{align}
\mathcal L(\theta_{0:t})
={}&\frac\lambda2\|\theta_0\|^2
+\frac12\sum_{i=1}^t(v_i-a_i^\top\theta_i)^2\nonumber\\
&+\frac\lambda2\sum_{i=1}^t\frac{\|\theta_i-\delta_i\theta_{i-1}\|^2}{1-\delta_i^2}.
\label{eq:trajectory}
\end{align}
The final component of the minimizer equals the filtered posterior mean at time $t$. This objective shows what selective forgetting changes: it permits the regression function itself to evolve, with a step-dependent penalty. In the unit-retention limit, the trajectory becomes constant and the objective reduces to ordinary ridge regression on the accumulated features. In the zero-retention limit, the recurrent regression state uses only the current observation; the local convolution still supplies short-range context.

\section{Block Maps and Their Adjoints}
\label{app:block}
\subsection{Constructing the covariance map}
Initialize $A=\I$, $C=0$, and $J=0$ for an empty block. For one observation with feature $a$, retention $\delta$, and prior variance $r=\lambda^{-1}$, form
\begin{align}
C^-&=\delta^2C+(1-\delta^2)r\I, & z&=C^-a,\nonumber\\
s&=1+a^\top z, & w&=\delta A^\top a.
\end{align}
The block statistics update by
\begin{equation}
A\leftarrow\delta A-zw^\top/s,\qquad
C\leftarrow C^--zz^\top/s,\qquad
J\leftarrow J+ww^\top/s.
\end{equation}
Substitution into Equation~\eqref{eq:riccati-map} gives the covariance produced by the token recurrence for any incoming positive definite $P$. The Woodbury identity gives Equation~\eqref{eq:riccati-compose}. Composition is associative because each triple represents a function of the incoming covariance. Applying prefix maps to $P_0$ therefore yields the same boundary states as sequential filtering.

\subsection{Causal source statistics}
Within a block, write $\ell_i=\sum_{j\leq i}\log\delta_j$. The coefficient transport and cross statistic at position $i$ satisfy
\begin{equation}
F_i=e^{\ell_i}\I-\sum_{j\leq i}e^{\ell_i-\ell_j}G_jw_j^\top,
\qquad E_i=\sum_{j\leq i}w_je_j^\top/s_j.
\end{equation}
For an output adjoint $\bar o_i=\partial\mathcal L/\partial o_i$, define $f_i=F_i^\top b_i$ and $z_i=E_i\bar o_i$. Then
\begin{equation}
D_{\rm local}=\sum_i f_i\bar o_i^\top,\qquad
S_{\rm local}=\sum_i(f_i z_i^\top+z_i f_i^\top).
\end{equation}
Each sum can be expressed using block Gram products and inclusive causal masks. These statistics combine with the terminal adjoints according to Equation~\eqref{eq:joint-adjoint}.

\subsection{The local handwritten backward recurrence}
At a token, add $b_i\bar o_i^\top$ to $D$, and set $d_G=De_i$ and $d_e=D^\top G_i$. With $z_i=\bar P_i a_i$ as in Equation~\eqref{eq:update}, compute
\begin{align}
d_s&=\frac{\tfrac12z_i^\top S z_i-d_G^\top z_i}{s_i^2}, &
\zeta_i&=\frac{d_G-Sz_i}{s_i}+d_s a_i,\nonumber\\
S^-&=S+\zeta_i a_i^\top+a_i\zeta_i^\top, &
D^-&=D-a_id_e^\top.
\end{align}
The state adjoints propagate as $S\leftarrow\delta_i^2S^-$ and $D\leftarrow\delta_iD^-$. The value gradient is $d_e$, and the remaining input gradients are
\begin{align}
\bar b_i&=\Theta_i\bar o_i,&
\bar a_i&=d_s z_i+\bar P_i\zeta_i-\bar\Theta_i d_e,\nonumber\\
\bar g_i&=\delta_i^2\tr\!\left[S^-(P_{i-1}-r\I)\right]
+\delta_i\tr\!\left[(D^-)^\top\Theta_{i-1}\right], & g_i&=\log\delta_i.
\end{align}
The prior receives $\tfrac12(1-\delta_i^2)\tr(S^-)$ per transition and $\tfrac12\tr(S_0)$ from the initial covariance. Block products evaluate the contractions with forward states from their saved boundaries and rank-one updates. These expressions propagate gradients through the complete training window.

\subsection{Inverse residual correction}
For $W=\I+C_lJ_r$, the eigenvalues are positive because $C_l,J_r\succeq0$. A spectral upper bound is
\begin{equation}
\lambda_{\max}(C_lJ_r)\leq
\min\!\left\{\|C_lJ_r\|_\infty,\tr(C_lJ_r),
\sqrt{\tr((C_lJ_r)^2)}\right\}.
\end{equation}
This bounds the interval used by a polynomial inverse approximation. Recomputing $R=\I-WX$ in working precision and updating $X^+=X+XR$ gives $\I-WX^+=R^2$ in exact arithmetic. The block recurrence uses two such residual corrections.

\begin{algorithm}[t]
\caption{Sequential Kernel Ridge Attention, one head}
\label{alg:kra}
\begin{algorithmic}[1]
\Require Features $a_t,b_t\in\R^D$, values $v_t\in\R^{d_v}$, retention $\delta_t$, ridge $\lambda>0$
\State $P\gets\lambda^{-1}\I_D$, $\Theta\gets0_{D\times d_v}$
\For{$t=1,\ldots,T$}
  \State $\bar\Theta\gets\delta_t\Theta$
  \State $\bar P\gets\delta_t^2P+(1-\delta_t^2)\lambda^{-1}\I_D$
  \State $z\gets\bar P a_t$, \quad $s\gets1+a_t^\top z$
  \State $e\gets v_t-\bar\Theta^\top a_t$
  \State $\Theta\gets\bar\Theta+(z/s)e^\top$
  \State $P\gets\bar P-zz^\top/s$
  \State $o_t\gets\Theta^\top b_t$
\EndFor
\State \Return $o_{1:T},P,\Theta$
\end{algorithmic}
\end{algorithm}

\section{Training and Reproduction Details}
\label{app:setup}
\subsection{Data and common training protocol}
We use the archived SlimPajama repository at revision \texttt{c60a39f1a375} and the GPT-2 tokenizer at revision \texttt{607a30d783df}; the accompanying manifest records both full revisions. A fixed permutation of source shard names determines reading order, and rows retain their source order. The SHA256 hash of each original UTF-8 document, interpreted as an integer modulo 1,000, assigns residues 0--9 to validation and all other residues to training. The resulting streams contain 92,783 training documents and 941 validation documents, with zero overlap in document hashes.

We append an end-of-text separator and pack contiguous streams into 2,049-token windows with stride 2,048, using the last 2,048 tokens as shifted labels. Each window resets the recurrent state; separators within a window preserve model state. The training stream contains 100,007,936 prediction tokens, and validation contains 1,048,576. The run manifest records the exact token-stream and index hashes. The reported validation stream serves as the shared development split for model construction and evaluation.

\begin{table}[ht]
\centering
\caption{Shared architecture and optimization settings.}
\label{tab:protocol}
\small
\begin{tabular}{ll@{\qquad}ll}
\toprule
Setting & Value & Setting & Value\\
\midrule
Layers / width & 6 / 384 & Heads / head width & 6 / 64\\
FFN width & 1,152 & Vocabulary & 50,257\\
Sequence length & 2,048 & Global / microbatch & 16 / 2\\
Updates & 3,052 & Devices & $8\times$ RTX 5090\\
Peak / final LR & $10^{-3}$ / $10^{-4}$ & Warmup updates & 100\\
Adam $(\beta_1,\beta_2)$ & $(0.9,0.98)$ & Adam $\epsilon$ & $10^{-8}$\\
Weight decay & 0.1 & Gradient clip & 1.0\\
RMSNorm $\epsilon$ & $10^{-6}$ & Logit softcap & 30\\
Parameter init. std. & 0.02 & Feature dimension & 64\\
Frequency count & 32 & Initial $\beta$ / $\lambda$ & $1/8$ / 1\\
\bottomrule
\end{tabular}
\end{table}

The first warmup learning rate is $10^{-6}$. Output and FFN down-projections use residual scaling $1/\sqrt{2L}$. Dropout is zero. The language-model head applies the shared logit softcap $30\tanh(\ell/30)$. Weight decay excludes the tied embedding, tensors of dimension below two, and explicitly exempt decay parameters. Parameters, gradients, and optimizer state are FP32; ordinary network activations use BF16 autocast. All models backpropagate through complete 2,048-token windows.

Model, data, bank, and numerical-check seeds are respectively 20261109, 20261110, 20261111, and 20261112. Common embeddings, projections, normalizations, and FFN parameters are loaded from a shared initial snapshot. Model-specific parameters use separate random streams. The content clock starts from the same scalar decay-gate initialization as the Gated DeltaNet baseline. On the fixed initial diagnostic inputs, the median head half-life, $\ln2/\operatorname{mean}(\rho)$, is 14.37 tokens. The clock remains trainable throughout.

\subsection{Backbones and recurrence controls}
Transformer++ uses causal softmax multi-head attention in the common residual backbone. Gated DeltaNet uses width-four Q/K/V convolutions, SiLU, normalized query/key vectors, and its native head normalization and output gate. The KDA and GDN-2 controls retain this outer layer and change the recurrence. Writing states in feature-by-value orientation, their updates are
\begin{align}
S_t^{\mathrm{KDA}}&=(\I-\eta_t k_tk_t^\top)\diag(\alpha_t)S_{t-1}
+\eta_t k_tv_t^\top,\\
S_t^{\mathrm{GDN2}}&=(\I-k_t(b_t^{\rm erase}\odot k_t)^\top)\diag(\alpha_t)S_{t-1}
+k_t(w_t^{\rm write}\odot v_t)^\top.
\end{align}
Expanded decay channels initially copy the scalar gate. Erase and write projections in the GDN-2 control begin equal and subsequently learn independently. Gates use the sigmoid range. The comparisons use direct gate projections in the common recurrent backbone.

The recurrent baselines use Flash Linear Attention \citep{fla2024}, pinned to commit \texttt{98193113ecb7} (full revision in the manifest). Gated DeltaNet and the two controls use chunk size 64 for training and recurrent kernels for decode. Recurrence gradients are checked against FP64 references. KRA uses the 64-token block-map and explicit-adjoint formulation of Section~\ref{sec:engineering}. The environment uses PyTorch 2.9.1 with CUDA 12.8 and Triton 3.5.1.

\subsection{Ablation controls}
The SiLU-versus-identity comparison uses the same post-projection convolution and initial tensors in both arms. The Fourier-bank pair shares pre-projection Q/K convolutions, SiLU content features, and a linear value path with post-projection convolution. Both arms have the same initial function; the learned-bank arm optimizes the spectral directions. The value-activation pair shares pre-projection convolutions on all three paths and compares SiLU values with linear values. Positive bandwidth and ridge remain learned across these comparisons. Table~\ref{tab:coreablations} separates the three groups and identifies the common front end for each contrast.

\subsection{Timing and state accounting}
The training-latency benchmark uses the final checkpoints and the shared sequence length and microbatch. Both models use the same normalization, feed-forward, and loss optimizations, plus captured execution. Gated DeltaNet also uses its native fused convolutions, gates, and output normalization. Full-model forward--backward latency is the median of 50 samples after five warmup iterations. The eight-GPU step includes distributed gradient reduction, gradient clipping, and AdamW. After capture warmup, five steps are discarded and 50 maximum-rank wall times are retained. Synchronization barriers lie outside the timed interval. A zero learning rate preserves each checkpoint during timing; every rank's parameters are checked afterward.

Persistent-state measurements use batch one after 4K and 8K prefixes. They include the recurrent matrices and convolution history, or the attention key/value cache. The accompanying numerical data contains the measurements for each model.

The six-layer KRA core stores
\begin{equation}
6\times6\times(64^2+64\times64)\times4
=1{,}179{,}648\ \text{bytes}.
\end{equation}
Its three short-convolution histories add 41,472 bytes, yielding 1,221,120 bytes, or 1.164551 MiB. The Fourier bank is a learned parameter and contributes no additional recurrent buffer. This storage is independent of prefix length. Training activations and saved boundary states scale with the training sequence length.

\section{Recall Definitions}
\label{app:recall}
Natural recall annotations identify the nearest preceding occurrence of an exact token bigram in the same document and window. A rare bigram has training frequency at most twelve. Accuracy predicts the second token from the current prefix; NLL uses the full vocabulary. Distances are measured between the matched occurrences. Annotations reset at the end-of-text separator, independently of the model-state reset convention.

For controlled recall, each of 128 cases supplies a 32-token natural cue, a distractor cue, a background, and two answer tokens $A,B$. For each distance $d\in\{64,128,256,512,1024,1536,1984\}$, we insert an earlier cue followed by either $A$ or $B$, and end the prompt with the query cue. The earlier cue either matches the query or is a distractor. This gives $128\times7\times2\times2=3,584$ prompts. Let $\ell_{c,s}(y)$ be the log probability of target $y$ with cue condition $c$ and stored answer $s$. Define
\begin{equation}
M_c=\frac12\left[\ell_{c,A}(A)-\ell_{c,B}(A)
+\ell_{c,B}(B)-\ell_{c,A}(B)\right],\qquad
\mathrm{Gain}=M_{\rm matched}-M_{\rm distractor}.
\end{equation}
Balanced accuracy compares the two target probabilities for each stored answer under matched cues, averages the two directions, and awards half credit for ties. It is a two-choice score; natural top-1 uses the full vocabulary. ``Far'' copy results average the four distances at least 512. This custom diagnostic measures association strength and answer choice separately.

\section{Numerical Validation}
We validate recurrent outputs, covariance and coefficient states, all input gradients, the learned prior, and complete model gradients. Eleven core cases cover non-diagonal positive definite initial covariance, tail blocks, long zero/weak forgetting, and captured 2K inputs from the first and last model layers. Six additional cases cover uniform strong and zero forgetting. Seven checks verify the shared prior's initial-state and transition contributions. Twenty-eight model checks cover FP32/BF16 execution, every learned parameter, causal suffix invariance, state capacity, and uneven streaming continuation.

An independent FP64 check verifies the recurrent--dual identity, covariance-map composition, causal block read, and joint adjoint formulas across 60 cases. It covers unit retention, near-unit retention, nonuniform content steps, strong forgetting, and tail blocks. The optimized complete model has a maximum parameter-gradient relative error of $7.47\times10^{-6}$ in the FP32 comparison. Its million-token NLL differs from the reference by $2.18\times10^{-6}$. The accompanying manifest records source hashes, checkpoint hashes, formula-check results, and the data used to generate the figures and tables.

\end{document}